# Adaptability of Computer Vision at the Tactical Edge: Addressing Environmental Uncertainty

Hayden Moore
*Carnegie Mellon University,
Software Engineering Institute,
AI Division*

hmmoore@sei.cmu.edu

**Abstract**

Computer Vision (CV) systems are increasingly being adopted into Command and Control (C2) systems to improve intelligence analysis on the battlefield, the tactical edge. CV systems leverage Artificial Intelligence (AI) algorithms to help visualize and interpret the environment, enhancing situational awareness. However, the adaptability of CV systems at the tactical edge remains challenging due to rapidly changing environments and objects which can confuse the deployed models. A CV model leveraged in this environment can become uncertain in its predictions, as the environment and the objects existing in the environment begin to change. Additionally, mission objectives can rapidly change leading to adjustments in technology, camera angles, and image resolutions. All of which can negatively affect the performance of and potentially introduce uncertainty into the system. When the training environment and/or technology differs from the deployment environment, CV models can perform unexpectedly. Unfortunately, most scenarios at the tactical edge do not incorporate Uncertainty Quantification (UQ) into their deployed C2 and CV systems. This concept paper explores the idea of synchronizing robust data operations and model fine-tuning driven by UQ all at the tactical edge. Specifically, curating datasets and training child models based on the residuals of predictions, using these child models to calculate prediction intervals (PI), and then using these PI to calibrate the deployed models. By incorporating UQ into the core operations surrounding C2 and CV systems at the tactical edge, we can help drive purposeful adaptability on the battlefield.

## 1 Introduction

Environmental uncertainty, defined as *"the degree to which future states of the world cannot be anticipated and accurately predicted"* [1], can limit a Command and Control (C2) system's ability to help battle commands plan, prepare, and perform different objectives in a rapid and organized fashion. The tactical edge can be defined as *"The platforms, sites, and personnel operating at lethal risk in a battle space with a strong dependency on information systems and operational readiness"* [2]. Here the environments, missions, and objectives can all quickly change, and potentially introduce uncertainty into the warfighter's C2 systems. C2 systems leverage computer vision (CV) to provide a more holistic *view* of the changing environments at the tactical edge. Unfortunately, CV models are data-driven and may suffer from large extrapolation errors when applied to changing objects and environmental conditions [3]. In other words, as the environments and objects existing in these environments begin to change (even slightly) the C2 and cascading CV systems may become miscalibrated and inaccurate. CV systems classify, predict, and localize different mission-specific objects and agents with some degree of confidence. Warfighters having confidence and accuracy versus unpredictability in their systems at the tactical edge can be the difference between life and death. Uncertainty quantification (UQ) is used to ensure model trustworthiness and improve warfighter's understanding of data limits and model deficiencies. This paper explores the idea of warfighters using UQ to influence the C2 and cascading CV systems. Specifically, calculating confidences and prediction intervals (PI), detecting data that is out-of-distribution (OOD), and curating relevant datasets to recalibrate the deployed model. Ultimately, warfighters can use UQ to help drive adaptability and promote robust and informative AI systems [4].

According to the Defense Advanced Research Projects Agency (DARPA), complex physical systems, devices, and processes important to the Department of Defense (DoD) are often poorly understood due to uncertainty in models, parameters, operating environments, and measurements [5]. Thus, given this defined complexity, warfighters should aim to create a battle rhythm that incorporates measuring the uncertainty of their CV systems. We want the warfighters operating at the tactical edge to have a stronger understanding of their deployed CV model's performance. Warfighters can use this measured uncertainty to directly influence future C2 and CV systems/operations. Again, this will allow for quicker adaptability in changing environments, improving the battle command's situational awareness.

## 1.1 Computer Vision at the Tactical Edge

CV can be defined as a specific AI system that enables computers to interpret visual information. It usually involves parsing visual data through algorithms, like Convolutional Neural Networks (CNNs), to detect, classify, and localize objects of interest. CV can provide visibility on a battlefield by constantly detecting objects in the surrounding environment. Warfighters can analyze the data captured from different edge sensors to provide actionable intelligence. CV can also help warfighters to see objects that are otherwise hidden or not clear to the naked eye.

CV models are trained with the intention of making accurate predictions on the objects and environments expected to be seen (or hidden) at the tactical edge. CV model training usually first involves curating a training dataset of visual data that represents the objects and environments expected to be seen. This data will be curated through different data operations like data collection, data labeling, data cleaning, and data transformation. All of these different data operations can be performed at the tactical edge and will allow the warfighters to effectively curate relevant data to be used to improve their models. The CV models will then attempt to learn the representations and distributions of the data passed to the model at training time.

CV models will also be validated and tested through a similar process. Datasets separate from the training set can be curated and used for validation and testing of the CV model. A test set can be curated for model testing and is to be exclusively used after the model has completed an entire training (or fine-tuning) cycle. The test dataset should be held out from training so that the test results can accurately reflect the model's ability to generalize to data it has never seen before. These different datasets are curated with the intention to improve and evaluate the model deployed at the tactical edge.

By using UQ, warfighters can more accurately measure where their CV model is failing and then begin to curate the necessary data and fine-tune the model. Figure 1 below shows how the data for a well-defined class can begin to change and degrade as time passes. This directly impacts the confidence in predictions from the model that is deployed at the tactical edge. This paper suggests that warfighters need to build UQ into their core CV operations. Warfighters should actively measure their deployed model's uncertainty, curate relevant datasets, fine-tune these models, and then redeploy these new models to the tactical edge.

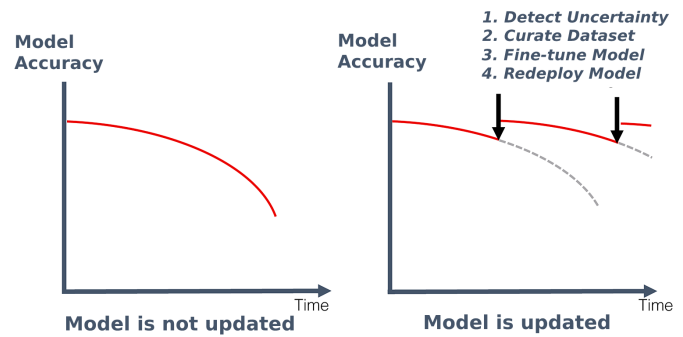

*Figure 1 - Data drift over time; Showing how well defined objects and environments can change as time passes. Eventually degrading the CV model's performance. [6]*

## 1.2 Improving warfighters Situational Awareness

Situational awareness can be defined as the process of perceiving the elements in the environment, understanding the elements in the environment, and the projection of their status into the near future [7]. C2 systems use situational awareness by *"exercising authority and direction by a properly designated commander over assigned and attached forces in the accomplishment of the mission"* [8]. The tactical edge can be dangerous and confusing and a thorough understanding of this environment will better prepare warfighters. Moreover, visibility is crucial on the battlefield and warfighters should leverage CV systems to gain a more robust situational awareness and visibility into their environment.

When we look at "The Situational Awareness Model" described by Mica Endsley, portions of a CV system operating within a C2 infrastructure can be imagined to sit at "Level 1". The model describes the functionality of the systems that operate at this level to be that of ones that help provide a *"perception of the elements in the current situation"* [9]. The deployed CV model, algorithms, and sensors/cameras all help to detect the objects and environments in the current situation. As stated before, CV enables computers to digest and interpret visual information, and in this context specifically, the elements and objects at the tactical edge. This perception of the objects then influences and supports the higher levels of this model. Eventually, it flows into the *Decision* and *Performance of Actions* stages at the tactical edge.

UQ can be thought to exist around the *Situational Awareness, Decision,* and *Performance of Action* phases. Eventually, this influences the *Feedback* phase of the model. The uncertainty measured from the CV system can be used to influence the proceeding *Decision* phase. Likewise, after a *Decision* and *Performance of Action* is conducted, UQ may be used to measure the confidence and overall effectiveness. Specifically, UQ will help to measure the precision of predictions and correctness of the CV model deployed at the tactical edge. These results will eventually flow into the *Feedback* phase of the model. Using this approach, UQ will create a more adaptable situational awareness for warfighters. A slightly modified diagram of this model is depicted in Figure 2.

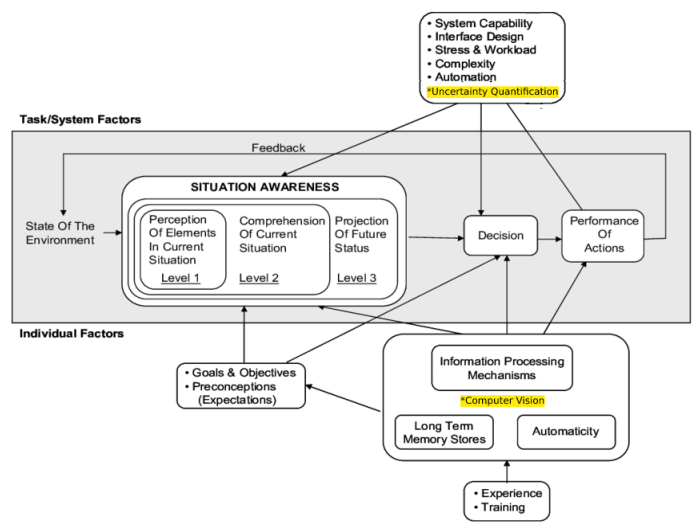

*Figure 2 - Modified Situational Awareness Model; Describing how situational awareness is measured, operated, and used for decisions on the battlefield. Modified to explicitly depict CV and UQ [10]*

## 2    Environmental Uncertainty

### 2.1    Unpredictable Environments at the Tactical Edge

Situational awareness, by its definition, requires a thorough understanding of environments and the objects within these environments. Unfortunately, the tactical edge presents rapidly shifting and unstable environments that are hard or sometimes impossible to predict. Also, the physical and environmental limitations of the battlefield present a unique challenge to those operating.

These limitations can include anything from the terrain and geography of the battlefield, the extreme weather conditions, to the changing objects that exist in these environments. Environmental factors such as high heat, cold, dust, wind, and rain can all affect a warfighter's ability to effectively perform their mission and may also obfuscate objects or confuse the model when making predictions.

Frequently changing environments make it difficult to properly train a model to make accurate predictions, as the model is trained to expect one thing but the tactical edge is presenting it differently, confusing the model. The physical environments and objects that are expected to be seen on the battlefield might not be well-defined or could be constantly changing. Models may be trained with data that anticipates a particular environment, shape, color, size, and so on. This makes it incredibly difficult to predict the exact environment and objects warfighters will encounter at the tactical edge. Further describing the necessity for frequently measuring uncertainty, collecting data, and fine-tuning the deployed model.

### 2.2    Differences in Controlled and Edge Environments

Typically, CV models used at the tactical edge will have their weights initialized and trained on data captured outside of the edge, in controlled development environments. Most likely, warfighters will use models that have already been trained on data captured outside of the edge. Then these models are transferred to the edge, where the model can be fine-tuned on data captured directly at the tactical edge. This process is also known as transfer learning, see [11], and allows for a model trained outside of a particular environment to be used as a starting point to then fine-tune the model to better represent the deployment environment. Transfer learning allows the model to maintain its performance achieved from the training outside of the deployment environment (controlled), and then calibrate the model to be more tailored for the deployment environment (tactical edge).

In simple terms, a controlled environment makes it easier to develop a *"good"* CV model since some of the environmental variables and objects are controllable to a certain degree. Understanding this, this data in controlled environments can introduce unintended biases to the model at training time. As stated before, the edge environment is hard to anticipate and the objects expected are not always well defined. These differences in the objects the model has seen at training time and the objects collected and observed at the edge can create uncertainty in predictions. CV models will be trained for a particular object, pattern, or environment at the tactical edge, and predictions will become more uncertain when changes are introduced and the data moves OOD.

These changes could be from introducing new technology, the environment drastically changing, the adversary being camouflaged, and so many other variations. These changes can affect the visual data captured and seen by the model at the tactical edge. Ultimately, these changes will create uncertainty in the model's predictions. This is because the model was trained on visual data that was captured with a specific technology, format, and representation. Meaning that to some degree the current CV model has learned the data distribution in which the training data was sampled. There is some bias in the model which will favor the visual data captured with the same technology, environment, and approach used at the model's training time. With this in mind, using UQ, warfighters can measure how their CV model has been affected by the changes at the tactical edge.

### 2.3 MEASURING UNCERTAINTY AT THE TACTICAL EDGE

Uncertainty can be hard to prepare for and even when it's quantified, fixing the uncertainty is not always trivial. Two types of uncertainty will typically impact CV and other AI systems, known as epistemic and aleatoric uncertainty. For this paper's scenario, epistemic uncertainty refers to the uncertainty caused by the lack of relevant and quality data captured at the tactical edge. In other words, this type of uncertainty is reducible to a certain degree, given that warfighters can collect the right data that represents their environment and fine-tune their model [12]. On the other hand, there exists aleatoric uncertainty and this refers to the inherent uncertainty that exists in these images captured. This is not the same as a poor-quality image or a technology change that causes some degree of uncertainty. Instead, aleatoric uncertainty describes the inherent stochasticity that will exist in these images regardless of the quality or technology the warfighters are using [12].

The following is an example where warfighters use uncertainty quantification to better understand their epistemic uncertainty in their models and datasets, eventually using this knowledge to fine-tune and improve their deployed models. This example was inspired by the PI3NN (Prediction Interval, using 3 Neural Networks) methodology featured in "Uncertainty Quantification of Machine Learning Models to Improve Streamflow Prediction Under Changing Climate and Environmental Conditions" from Oak Ridge National Laboratory [3].

We can imagine $f(x)$ as the foundational CV model trained for mission specific objects and environments at the tactical edge. When $f(x)$ is first introduced to this environment, the only data the model has seen is either from previous missions or controlled environments. Either way, the data that the model has seen most likely does not *exactly* reflect the environment and objects at the tactical edge. This can be described as the environmental bias and essentially is the difference between the ground truth and the models predictions. We want this difference to be minimized so that warfighters can maintain their tactical advantage. The given confidence, defined as γ, will represent the confidence across a specific section of data. The confidence will change based on the changing environment and objects, but will always fall in the bounds $\gamma \in [0, 1]$. To maintain a high confidence in predictions, warfighters should repeatedly conduct robust data operations, make predictions on their environment, and store data that is collected directly at the edge. When *ample* data has been accumulated into a dataset, defined as $D_{edge} = \{(x_i, y_i)\}_{i=1}^{N}$, warfighters can begin to look into the uncertainty with the intention to improve the deployed model.

The following approach uses three different CV models to help measure the uncertainty at the tactical edge. The first being $f(x)$ which represents the deployed model. Also, $u(x)$ and $l(x)$ will represent the upper and lower bound models of the deployed model. We can construct the datasets to train $u(x)$ and $l(x)$ where the curation is driven by the residuals, or uncertainty in predictions from $f(x)$. Using root-finding methods, we can determine the upper and lower bounds of the predictions for a narrow section of data. Using the parent dataset $D_{edge}$, an upper and lower bound dataset can be curated from the following,

$D_1 = \{(x_i, y_i - f(x_i)) | y_i >= f(x_i), i = 1, ..., N\}$,
$D_0 = \{(x_i, f(x_i) - y_i) | y_i < f(x_i), i = 1, ..., N\}$

[3]. $D_1$ contains the data points of positive residuals (model is overestimating) and $D_0$ is the data points of negative residuals (model is underestimating). Once we have curated $D_1$ and $D_0$, these new datasets can be used to help calculate uncertainty, and fine-tune the deployed model, eventually generating two new child models defined as $u(x)$ and $l(x)$. Now we will define new functions $U(x)$ and $L(x)$, defined as $U(x) = f(x) + \alpha u(x)$ and $L(x) = f(x) - \beta l(x)$ [3]. By making $u(x)$ and $l(x)$ a function of $f(x)$ and introducing two new coefficients α, β, we help ensure we

calculate the uncertainty in our deployed model for a narrow section of data [3], given our current confidence. α, β are calculated using a bisection method, a root-finding method used with polynomial equations.

$$Q_1(\alpha) = \sum_{(x_i, y_i) \in D_1} 1_{y_i > U(x_i)}(x_i, y_i) - \frac{N(1-\gamma)}{2},$$

$$Q_0(\beta) = \sum_{(x_i, y_i) \in D_0} 1_{y_i < L(x_i)}(x_i, y_i) - \frac{N(1-\gamma)}{2} \ [3].$$

Finally, we can calculate an accurate and precise prediction interval for a specific section of data given the current confidence at the tactical edge, defined as $[L(x), U(x)]$. This interval is equivalent to $[f(x) - \beta l(x), f(x) + \alpha u(x)]$ which is the exact same as $\gamma N$ [3]. This approach of UQ aims to produce a well-calibrated prediction interval for a specific section of data [3]. This can be used to help drive adaptability on the battlefield, showing the warfighters where their CV system is failing and how much confidence their model has across a specific section of data.

Referencing the results from "Uncertainty quantification of machine learning models to improve streamflow prediction under changing climate and environmental conditions", the researchers used this approach along with a Long short-term memory (LSTM) model. They were able to show improvements and reveal insights into prediction accuracy, high-quality predictive uncertainty quantification, and model robustness when applied to changing environments and weather conditions [3].

## 3    Improving Adaptability

### 3.1    Using Measured Uncertainty to Drive Adaptability

Ultimately, warfighters should use UQ to help drive decision making, changes in their C2 systems, and deployment of different operations. More specifically, warfighters can use the calculated interval $[L(x), U(x)]$ to drive the deployment and changes in robust data operations and model fine-tuning. As the CV models uncertainty is quantified, warfighters should deploy operations to directly help the AI system gain confidence in that specific area [4]. A high-level battle rhythm demonstrating how UQ can be built into these operations is represented in Figure 3.

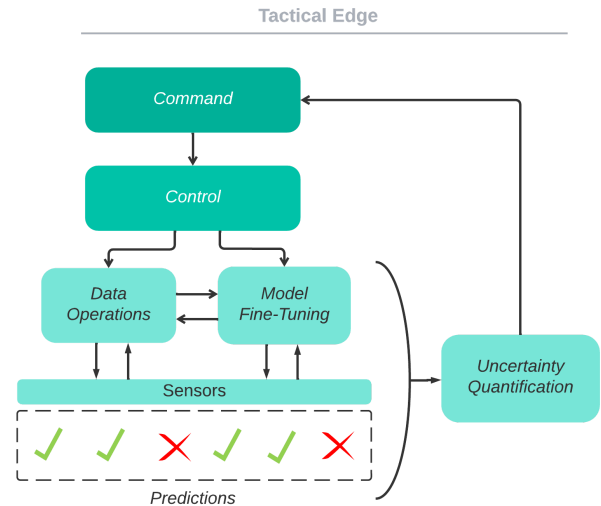

*Figure 3 - C2 and CV System using UQ; Showing how UQ can be used to feedback into the C2 system for better adaptability at the tactical edge.*

### 3.2    Data Operations and Model Fine-Tuning

Warfighters will use the calculated uncertainty to orchestrate data operations at the tactical edge. When warfighters detect uncertainty for a particular set of data, this information should be fed back into the C2 system. This will facilitate the warfighters to quickly deploy more robust data operations to help the model gain confidence in the measured area of uncertainty [3]. More specifically, the warfighters should deploy data collection and data curation efforts for that specific section of objects or environments showing high levels of uncertainty. Using newly collected data and the data used to calculate the uncertainty, warfighters can curate an optimized dataset to help improve the model's recently measured uncertainty.

An example could include a scenario where a CV model deployed at the edge begins to show uncertainty in its predictions for a particular enemy tank. Adversaries are constantly trying to gain the tactical advantage at the edge by changing their expected appearances, changing technology, and using camouflage to obfuscate detections. As warfighters begin to measure uncertainty in their CV system, specifically for this example, a particular enemy tank, the C2 system should be informed, and data operations and model fine-tuning should be deployed to help gain confidence. In this scenario, this means warfighters collecting and curating new datasets, and deploying data labeling efforts, for the enemy tank showing uncertainty. Using the data captured directly from the tactical edge (and labeling it accurately) creates a dataset that is more representative of the enemy tank in the context of the current environment.

Finally, this dataset can then be used to fine-tune the foundational model currently deployed.

By using UQ warfighters can see where their CV model is failing. UQ provides a metric that can be used to directly measure a model's uncertainty and performance within a narrow section of data. UQ provides more insight into CV models operating at the tactical edge and also directly influences future data operations and model fine-tuning efforts.

When warfighters *complete* a cycle of data curation tasks motivated by the recently measured uncertainty, they can begin to fine-tune the foundational model with this new data. Over time this newly fine-tuned model, defined as $f(x)_{t'}$ will become the new model deployed at the tactical edge. Where, $f(x)_{t'}$ was trained on the recently curated data from robust data operations, driven by the measured uncertainty.

### 3.3 Building a More Robust Tactical Edge

We can imagine this process as a constant battle rhythm that is frequently changing and adapting to the changing environment at the tactical edge. Creating a cycle of conducting robust data operations, fine-tuning the foundational model, and then deploying this new model back into the field, where these operations are driven by UQ. This cycle creates a CV model that is constantly evolving to best represent the tactical edge, providing warfighters with a more adaptable situational awareness.

In conclusion, this paper suggests that UQ should be integrated into the regular processes conducted at the tactical edge, having UQ be part of the fundamental pipeline. Warfighters can actively monitor the performance and uncertainty of their CV models deployed, using this uncertainty to feedback to their core C2 systems. Ultimately, creating a faster and more adaptable situational awareness for warfighters operating in the field. This paper aspires to conduct future experiments and gather empirical evidence to support how UQ improves the adaptability of C2 and CV systems in changing environments at the tactical edge.

*and methods - machine learning*. SpringerLink. https://link.springer.com/article/10.1007/s10994-021-05946-3